\documentclass[final,5p,times,twocolumn]{elsarticle}

\usepackage{amssymb}

\usepackage{graphicx}
\usepackage{tabularx}
\usepackage{color}
\usepackage{multirow}
\usepackage{amsmath}
\usepackage{amssymb}
\usepackage{subfigure}
\usepackage{enumitem}
\usepackage{hyperref}
\usepackage[export]{adjustbox}
\usepackage{booktabs,caption}
\usepackage[flushleft]{threeparttable}

\usepackage{array}
\usepackage{booktabs}
\makeatletter
\def\hlinewd#1{%
  \noalign{\ifnum0=`}\fi\hrule \@height #1 \futurelet
   \reserved@a\@xhline}
\makeatother

\journal{arXiv}

\begin{document}

\begin{frontmatter}



\title{Deep Multiple Instance Learning for Airplane Detection in High Resolution Imagery}


\author[IUST]{Mohammad Reza Mohammadi}
\ead{mrmohammadi@iust.ac.ir}

\address[IUST]{School of Computer Engineering, Iran University of Science and Technology, Tehran, Iran}


\begin{abstract}
Automatic airplane detection in aerial imagery has a variety of applications.
Two of the significant challenges in this task are variations in the scale and direction of the airplanes.
To solve these challenges, we present a rotation-and-scale invariant airplane proposal generator.
We call this generator symmetric line segments (SLS) that is developed based on the symmetric and regular boundaries of airplanes from the top view.
Then, the generated proposals are used to train a deep convolutional neural network for removing non-airplane proposals.
Since each airplane can have multiple SLS proposals, where some of them are not in the direction of the fuselage, we collect all proposals corresponding to one ground-truth as a positive bag and the others as the negative instances.
To have multiple instance deep learning, we modify the loss function of the network to learn from each positive bag at least one instance as well as all negative instances.
Finally, we employ non-maximum suppression to remove duplicate detections.
Our experiments on NWPU VHR-10 and DOTA datasets show that our method is a promising approach for automatic airplane detection in very high-resolution images.
Moreover, we estimate the direction of the airplanes using box-level annotations as an extra achievement.

\end{abstract}


\end{frontmatter}

\section{Introduction}

Remote sensing is a contactless technique of gathering information.
Since the beginning of earth observation from space, many satellites have launched into space, which has used successfully in a wide range of civil, agricultural, and military applications \cite{seelan2003remote,james2019below,khan2018modern,li2020deep,sizkouhi2020automatic}.
With the development of very high resolution (VHR) imaging equipment, the resolutions of the available images have increased in spatial, spectral, and temporal domains.
Given these extensive and valuable data, automatic analysis of VHR images has been increasingly interested.

Automatic object detection in VHR images is a key module in a wide range of applications and receives significant attention in recent years \cite{cheng2016survey}.
In this paper, we focus on airplane detection in VHR images.
Although airplane detection has studied since many years ago, it is still a challenging problem because of the cluttered background, appearance and shape variations, different resolutions of satellite images, and the arbitrary orientation of airplanes.

\begin{figure}[!t]
	\centering
	\includegraphics[width=0.75\linewidth]{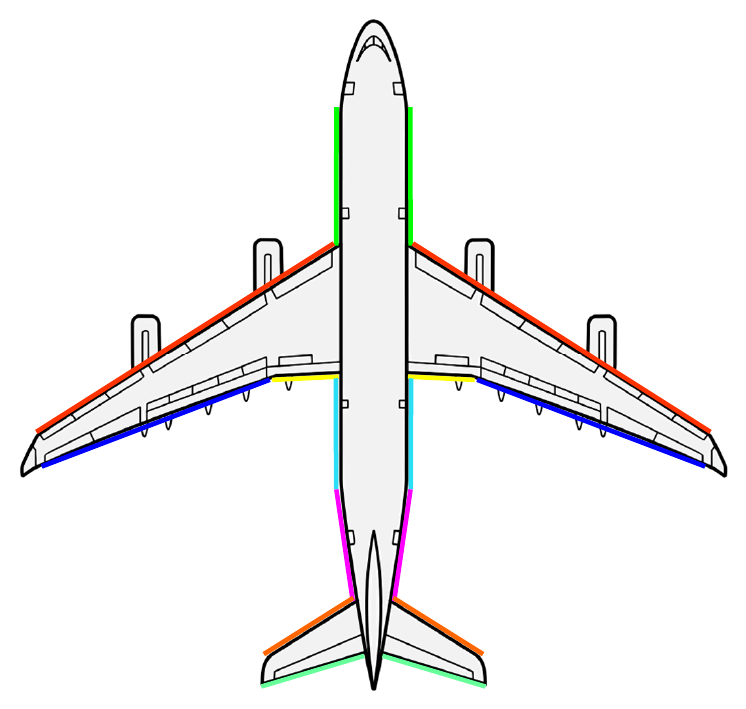}
	\caption{Outline drawing airplane in a flat style (top view)}
	\label{fig:Airplane}
\end{figure}

To obtain a rotation-and-scale invariant airplane detection algorithm, we employ the common characteristics of airplanes.
An airplane is a human-made object that is seen symmetrically from the top view (Fig. \ref{fig:Airplane}).
Also, the boundary of each airplane is a regular shape that can be approximated by a chain of line segments.
According to these characteristics, we develop an algorithm to generate airplane proposals called symmetric line segments (SLS).

Although the SLS proposal generator can detect almost all airplanes, there are many other regions in the VHR images contain symmetric line segments.
So, we need to refine them using an image classifier.
In recent years, Convolutional Neural Networks (CNNs) have achieved state-of-the-art results in several computer vision tasks \cite{krizhevsky2012imagenet,simonyan2014very,huang2017densely,james2019below}.
In this paper, we use CNN to classify the airplane candidates from the others.

Training a deep convolutional neural network requires a significant number of instances with the desired labels.
We can consider a proposal as a positive instance if its line segments have high overlap with the bounding box of an aircraft.
However, as can be seen in Fig. \ref{fig:Airplane}, some of the symmetric line segments on the airplane boundary are not symmetric with respect to the fuselage.
Therefore, we consider all proposals have high overlap with an airplane as a positive bag in which classifying one of them as positive is sufficient.
This problem calls multiple instance learning (MIL) \cite{amores2013multiple} in the machine learning literature.
In this paper, we propose a loss function to train the CNN with MIL formulation.

\begin{figure*}[!t]
	\centering
	\includegraphics[width=\linewidth]{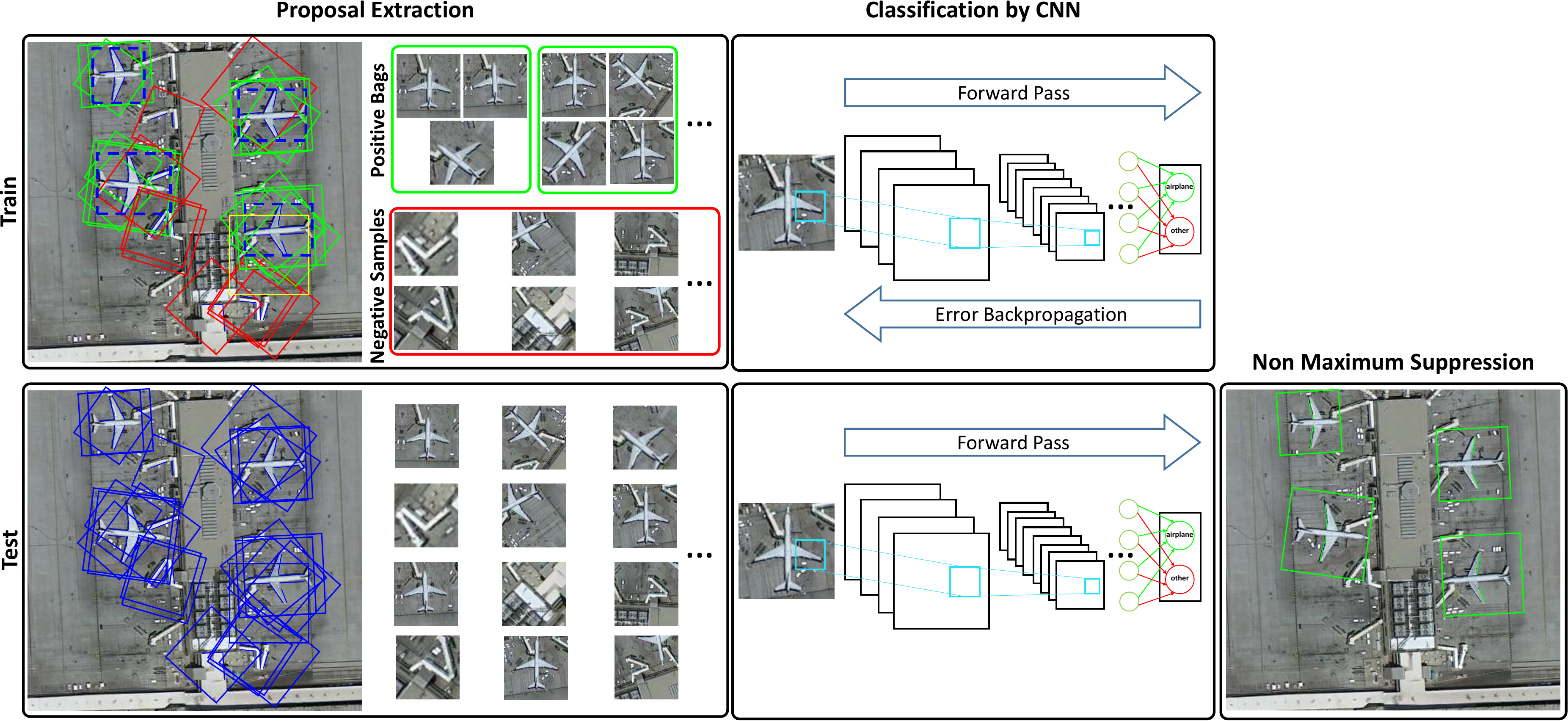}
	\caption{Schematic view of the proposed airplane detection approach}
	\label{fig:ProposedApproach}
\end{figure*}

In the test phase, we use non-maximum suppression (NMS) after CNN to eliminate the redundant detections.
The combination of the proposed deep multiple instance learning and non-maximum suppression leads to detect the most common SLS among airplanes.
We will show in the experiments that the most common SLS among airplanes have formed from two line segments of two wings that are symmetrical about the fuselage.
As a result, we can estimate the direction of the airplanes using box-level annotations.
The schematic view of the proposed approach is shown in Fig. \ref{fig:ProposedApproach}.

The contributions of this paper can be summarized as follows:
1) We introduce a novel proposal generation algorithm for airplanes called symmetric line segments (SLS).
2) We define a multiple instance learning based loss function to train the deep convolutional neural network.
3) We estimate the airplane direction using box-level annotations.
4) We validate our framework on NWPU VHR-10 \cite{cheng2014multi} and DOTA \cite{xia2018dota} datasets.

The rest of this paper is organized as follows.
Section \ref{sec:RelatedWorks} gives a brief overview of the existing airplane detection algorithms.
In Section \ref{sec:ProposedApproach}, we present our proposed approach.
We report and analyze the experimental results on NWPU VHR-10 and DOTA datasets in Section \ref{sec:ExperimentalValidations}, and finally, conclude the paper in Section \ref{sec:Conclusion}.

\section{Related Works}
\label{sec:RelatedWorks}

Generally, object detection algorithms consist of three main modules: proposal generation, feature extraction, and classification.
We discuss these modules as follows.

\subsection{Proposal Generation}
\label{sec:RelatedWorksProposalGeneration}

Proposal generation is one of the main differences between the published object detection algorithms, which its accuracy directly affects the next steps.
The most straightforward proposal generator is the sliding window that has been used in many studies, such as \cite{sun2012automatic,redmon2016you,liu2016ssd}.
A sliding window is a rectangular region of fixed size that slides across an image.
The number of generated proposals by this approach is very high, especially if we want to support different scales and aspect ratios.

To reduce the number of generated proposals and to make them more meaningful, the researchers have proposed several alternative algorithms.
Bo and Jing \cite{bo2010region} used image segmentation to combine neighbor pixels and generate homogeneous regions as the proposals.
To generate multi-scale proposals, Li et al. \cite{li2012automatic} employed multiple segmentations.
Moreover, some generic proposal generators have been developed in recent years, such as EdgeBoxes \cite{zitnick2014edge}, Selective Search \cite{uijlings2013selective}, BING \cite{cheng2014bing}, and RIGOR \cite{humayun2014rigor}.
For a more in-depth survey of generic proposal generators, we refer the readers to \cite{hosang2016makes}.
Many of the recent airplane detection studies have used these generic proposal generators.
EdgeBoxes has been used in \cite{khan2017automatic,farooq2017efficient,li2018aircraft}, BING in \cite{wu2015fast,luo2016airplane,zheng2016object}, and Selective search in \cite{long2017accurate}.

Although the generic proposal generators are also useful for airplane detection, the particular geometry of aircraft has led to the development of some specific proposal generators.
Liu et al. \cite{liu2020aircraft} proposed to use corner clustering for aircraft proposal generation.
Kawato et al. \cite{kawato2001circle} introduced the well-known circle-frequency filter (CFF), which has been used in many works such as \cite{cai2006airplane,gao2013aircraft,an2014automated,zhang2015unsupervised}.
CFF developed based on the cross shape of the airplanes.
More precisely, the intensities along a circle with a proper radius and an appropriate center on the airplane will change regularly from darkness to brightness and will repeat four times.
To detect multi-scale airplanes, one needs to use CFF with different radii.
In this paper, we propose a specific airplane proposal generator based on the symmetric line segments.

\subsection{Feature Extraction}
\label{sec:RelatedWorksFeatureExtraction}

The number of false alarms produced by a proposal generator, even the specific ones, is usually much more than acceptable.
Thus, we require to eliminate the undesired proposals by a supervised algorithm.
In the feature extraction module, we represent each proposal by a discriminative feature vector.
Researchers have developed various geometrical and textural feature extraction methods, some of which are reviewed here.

The boundary of an airplane in the top view is quite distinguished from other objects.
So, simple shape descriptors can be useful features.
Hosomura et al. \cite{hosomura2010airplane} extracted some simple geometrical features such as area, perimeter, roundness, and aspect ratio from the boundaries.
In another study, Inglada \cite{inglada2007automatic} used higher-level geometrical features.
However, the performance of these features is very sensitive to the accuracy of boundary extraction that is a challenging problem yet.

Textural features are used more frequently than geometrical ones in this field.
Histogram of oriented gradients (HOG) was introduced in \cite{dalal2005histograms} and have been used in many airplane detection studies such as \cite{an2014automated,cheng2013object,cheng2016object}.
Bag-of-words (BoW) is another useful feature extractor, which has been employed by Bai et al. \cite{bai2014vhr} for airplane detection.
Gabor filters, Local Binary Patterns (LBP), and Haar-like features are some other conventional feature extractors that Kumar and Bhatia surveyed them in \cite{kumar2014detailed}.

Despite the progress made in the design of engineered features, their performance has saturated in recent years.
On the opposite side, feature learning algorithms have been being more popular every day.
Feature learning is a set of methods to automatically discover the representations required for feature classification from raw data.
Convolutional neural networks \cite{lecun1998gradient,szegedy2015going,howard2017mobilenets,tan2019efficientnet} are among the most successful feature learning algorithms from images that have been used in \cite{zhong2018multi,deng2018multi,zou2018random,liu2020aircraft} for airplane detection.
In this paper, we employ a powerful CNN for feature extraction.

\subsection{Classification}
\label{sec:RelatedWorksClassification}

The last module in a typical object detector pipeline is decision making by a classifier.
SVM and AdaBoost are two popular classifiers that have been used in \cite{li2011saliency,cheng2013object,bai2014vhr,cheng2014scalable,cheng2016learning} and \cite{an2014automated,petridis2008learning}, respectively, for airplane detection.
These classifiers are commonly used after engineered features.
Although such classifiers can also be used after learned features, a multi-layer perceptron is typically employed for this purpose.
The ability of end-to-end training of the feature extraction and classification modules is the main reason for using the MLP classifier after the feature learning module.
In this paper, we define a multiple instance learning based loss function and use a simple neural network for classification.

\section{Proposed Approach}
\label{sec:ProposedApproach}

The schematic view of the proposed approach is shown in Fig. \ref{fig:ProposedApproach}.
As can be observed, this approach consists of three modules discussed in the following sections.

\subsection{Proposal Generation}
\label{sec:ProposalGeneration}

\begin{figure}[!t]
	\centering
	\subfigure[]
	{
		\includegraphics[width=0.46\linewidth]{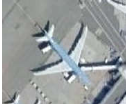}
		\label{fig:SLS_1}
	}
	\subfigure[]
	{
		\includegraphics[width=0.46\linewidth]{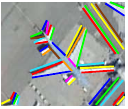}
		\label{fig:SLS_2}
	}
	\subfigure[]
	{
		\includegraphics[width=0.46\linewidth]{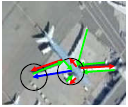}
		\label{fig:SLS_3}
	}
	\subfigure[]
	{
		\includegraphics[width=0.46\linewidth]{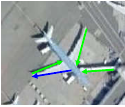}
		\label{fig:SLS_4}
	}
	\caption{Proposal generation based on the symmetric line segments, a) original image, b) line segments detected by LSD, c) line segments nearby a selected line segment, and d) line segments that are symmetrical enough to the reference line segment}
	\label{fig:SLS}
\end{figure}

Airplanes are human-made objects with a regular and symmetrical boundary from the top view.
We can approximate the airplane boundary by a chain of line segments.
Based on these properties, we propose a new airplane proposal generator called symmetric line segments (SLS).
The stages of SLS are shown in Fig. \ref{fig:SLS}.

Line detection is the first stage of SLS.
Hough transform \cite{duda1972use} is a widely used tool for line detection.
Hough transform does not consider the gradient direction, and so, it cannot detect small and noisy line segments.
On the other hand, line segment detector (LSD) \cite{von2010lsd} is a newer tool that has considered the gradient direction.
In Fig. \ref{fig:SLS_2}, we draw the line segments detected by LSD.

In the next stage, we select pairs of line-segments that can belong to an airplane.
Two line-segments of an appropriate proposal should have the following characteristics:
1) their endpoints have to be close together, and 2) they must be symmetric with respect to the central axis of the airplane (i.e., fuselage direction).
Thus, for a candidate line-segment such as the blue line-segment in Fig. \ref{fig:SLS_3}, we first select other line segments that at least one of their endpoints are in a limited distance from the endpoints of the candidate line-segment.
Then, we omit the pairs that are not sufficiently symmetrical.
We quantify the symmetry of two line-segments as follows.

\begin{figure}[!t]
	\centering
	\includegraphics[width=0.40\linewidth]{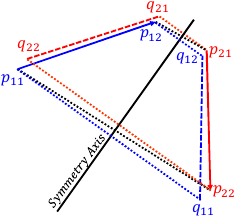}
	\caption{Measuring the symmetry degree of two line-segments}
	\label{fig:Symmetry}
\end{figure}

\begin{figure}[!t]
	\centering
	\includegraphics[width=1.0\linewidth]{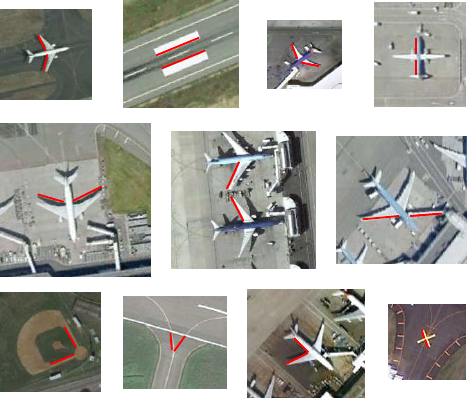}
	\caption{Sample proposals generated by the SLS method}
	\label{fig:Proposals}
\end{figure}

Let us consider \(p_{11}\) and \(p_{12}\) as the endpoints of the first line-segment, and similarly \(p_{21}\) and \(p_{22}\) for the second one.
An essential property of LSD is specifying the line-segment direction based on the gradient direction.
Thus, for an appropriate pair, \(p_{11}\) and \(p_{12}\) should be the mirrors of \(p_{22}\) and \(p_{21}\), respectively.
We define the line that passes from the points \(\left(p_{11}+p_{22}\right)/2\) and \(\left(p_{12}+p_{21}\right)/2\) as the symmetry axis for these line-segments.
Then, we mirror the line-segments around the symmetry axis and call the resulted endpoints \(q_{11}\), \(q_{12}\), \(q_{21}\), and \(q_{22}\), respectively.
A symbolic display of this concept is shown in Fig. \ref{fig:Symmetry}.
Finally, we compute the relative euclidean distance of the original and the mirrored endpoints as the measure of symmetry:

\begin{align}
	sym &= \frac{\left\|p_{11} - q_{22}\right\|_2 + \left\|p_{12} - q_{21}\right\|_2}{\left\|p_{11} - p_{22}\right\|_2 + \left\|p_{12} - p_{21}\right\|_2} \nonumber \\
	    &= \frac{\left\|q_{11} - p_{22}\right\|_2 + \left\|q_{12} - p_{21}\right\|_2}{\left\|p_{11} - p_{22}\right\|_2 + \left\|p_{12} - p_{21}\right\|_2}
\end{align}

where \(\left\|\:\cdot\:\right\|_2\) is the L2 norm (Euclidean distance).
We select the pairs with \(sym < 0.3\) as the airplane proposals (Fig. \ref{fig:SLS_4}).
As can be observed, one line segment may be present in more than one proposal or none of them.
This proposal generator is rotation-and-scale invariant; so, we can use it effectively in aerial imagery with different resolutions.
Some positive and negative proposals are shown in Fig. \ref{fig:Proposals}.

\subsection{Feature Extraction and Classification}
\label{sec:FeatureExtractionAndClassification}

\begin{figure}[!t]
	\centering
	\includegraphics[width=1.0\linewidth]{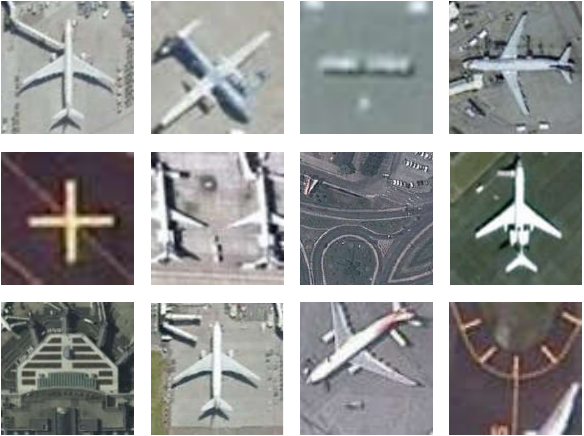}
	\caption{Sample cropped and resized proposals}
	\label{fig:Crops}
\end{figure}

\begin{figure}[!t]
	\centering
	\subfigure[]
	{
		\includegraphics[width=0.22\linewidth]{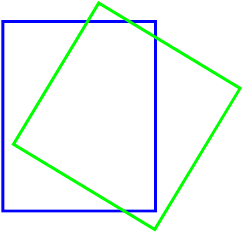}
		\label{fig:IoU1}
	}
	\subfigure[]
	{
		\includegraphics[width=0.22\linewidth]{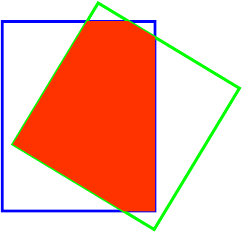}
		\label{fig:IoU2}
	}
	\subfigure[]
	{
		\includegraphics[width=0.22\linewidth]{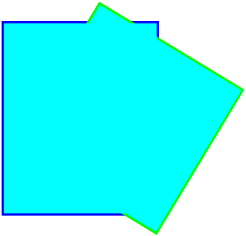}
		\label{fig:IoU3}
	}
	\caption{Computation of intersection over union (IoU)}
	\label{fig:IoU}
\end{figure}

To extract rotation-and-scale invariant representations from the SLS proposals, we crop a rotated square for each proposal with the following parameters: 

\begin{align}
	& x = \frac{x_{11} + x_{12} + x_{21} + x_{22}}{4} \\
	& y = \frac{y_{11} + y_{12} + y_{21} + y_{22}}{4} \\
	& side = 1.5 \times \max\left(\left\|p_{11}-p_{21}\right\|_2, \left\|p_{11}-p_{22}\right\|_2,\right.\\ \nonumber
	& \ \ \ \ \ \ \ \ \ \ \ \ \ \ \ \ \ \ \ \ \ \ \ \left.\left\|p_{12}-p_{21}\right\|_2, \left\|p_{12}-p_{22}\right\|_2\right)\\
	& \theta = atan2\left(y_{12} + y_{21} - y_{11} - y_{22}, x_{12} + x_{21} - x_{11} - x_{22}\right)
\end{align}

where \(p_{11}=\left(x_{11},y_{11}\right)\), \(p_{12}=\left(x_{12},y_{12}\right)\), \(p_{21}=\left(x_{21},y_{21}\right)\), and \(p_{22}=\left(x_{22},y_{22}\right)\) are the endpoints of the line segments.
Then, we resize each crop to the fixed size of \(64\times 64\) pixels that some samples are depicted in Fig. \ref{fig:Crops}.
To train the CNN, one needs to assign a label to each proposal.
We use the intersection over union (IoU) measure that is shown in Fig. \ref{fig:IoU}.
Intersection and union regions of the rectangles of Fig. \ref{fig:IoU1} are shown in Fig. \ref{fig:IoU2} and Fig. \ref{fig:IoU3} with red and cyan colors, respectively.
We label each proposal based on the following intervals:

\begin{align}
	label = \left\{
	\begin{array}{rl}
		+1      &   \ \ \ \  \texttt{if}\ \ \texttt{IoU} >= 0.3 \\
		 0      &   \ \ \ \  \texttt{if}\ \ \texttt{IoU} < 0.2 \\
		-1      &   \ \ \ \  \texttt{if}\ \ \texttt{IoU} >= 0.2\ \&\ \texttt{IoU} < 0.3
	\end{array}
	\right.
\end{align}

\begin{figure}[!t]
	\centering
	\includegraphics[width=1.0\linewidth]{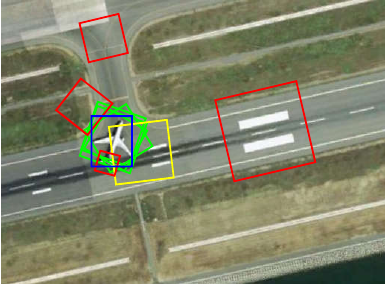}
	\caption{Labeling proposals based on the intersection over union with the ground truth bounding box (blue: ground truth, green: label +1, yellow: label -1, red: label 0)}
	\label{fig:Labeling}
\end{figure}

Some proposals with the corresponding labels are shown in Fig. \ref{fig:Labeling}.
In the training phase, we use the proposals with labels \(+1\) and \(0\) as the airplane and other objects, respectively, and ignore the proposals with label \(-1\).
As stated before and can be seen in Fig. \ref{fig:Labeling}, each airplane may have more than one corresponding proposal.
Moreover, due to the weakly labeling of the targets (box-level annotations), some of the proposals with label +1 may be undesirable in which the symmetry axis is not same as the main axis of the airplane.
Therefore, we let the classifier misclassify some of the positive proposals as follows.

We consider all proposals highly overlapped with one ground truth bounding box as a unique positive bag.
On the other hand, we consider all negative proposals as negative instances.
In each iteration, we update the weights of the network based on a batch, including some negative instances and some positive bags.
For the negative instances, we use the following focal loss:

\begin{align}
	loss_{neg}(p) = - p^3 log(1 - p)
\end{align}

where \(p \in [0\ 1]\) is the model’s estimated probability for the positive (airplane) class.
We use the focal loss defined in \cite{lin2018focal} with \(\gamma=3\) to down-weights the loss assigned to well-classified negative instances.
This loss focuses training on a sparse set of hard negative instances and prevents the vast number of easy ones from overwhelming the classifier.
On the other hand, for each positive bag, at least one instance should be classified as airplane.
Thus, we define the following loss for the instances of a positive bag:

\begin{align}
	& loss_{pos}(p) = \left\{
	\begin{array}{ll}
		 CE(p, 1)      &   \ \  \texttt{if}\ \ CE(p, 1) < 2 \times CE_{min}\\
		 0      &   \ \  \texttt{otherwise}
	\end{array}
	\right. \\
	& CE(p, 1) = - (1 - p) log(p)
\end{align}

where \(CE\) is the well-known cross-entropy loss, and \(CE_{min}\) is the minimum \(CE\) for the instances of this bag.
In other words, we ignore the instances with a relatively large loss.
By this approach, in the first iterations that the model is not converged, \(CE_{min}\) will be a large value, and subsequently, almost all instances in any positive bag are involved in the training.
While the model tends to converge, \(CE_{min}\) will decrease, and undesired positive instances will remove from the training.
As reported in section \ref{sec:ExperimentalValidations}, the trained model rejects about 55\% of the positive instances but less than 1\% of the positive bags.

\subsection{Non-Maximum Suppression}
\label{sec:NonMaximumSuppression}

\begin{figure*}[!t]
	\centering
	\subfigure[]
	{
		\includegraphics[width=.3\linewidth]{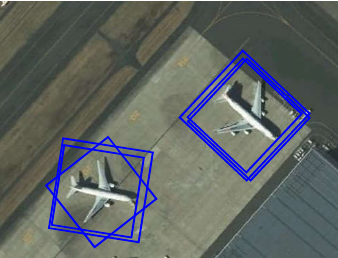}
		\label{fig:NMS1}
	}
	\subfigure[]
	{
		\includegraphics[width=.3\linewidth]{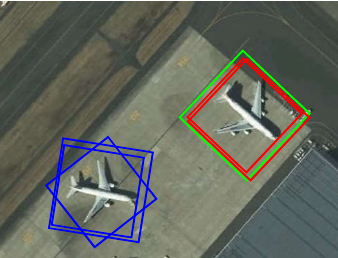}
		\label{fig:NMS2}
	}
	\subfigure[]
	{
		\includegraphics[width=.3\linewidth]{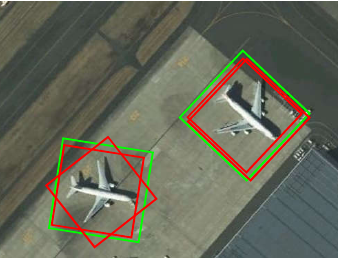}
		\label{fig:NMS3}
	}
	\caption{Non-maximum suppression procedure}
	\label{fig:NMS}
\end{figure*}

At test time, we estimate the score of each proposal using the trained network.
Given all scored proposals in an image, we apply a greedy non-maximum suppression (NMS).
In the implemented NMS, we iteratively select the highest score proposal and reject all proposals that have a significant overlap with this one.
This non-maximum suppression (NMS) approach is shown in Fig. \ref{fig:NMS}.
The detected squares by the network are drawn in Fig. \ref{fig:NMS1}.
The highest score proposal is drawn in Fig. \ref{fig:NMS2} with green color and the removed ones with red color.
The next iteration is depicted in Fig. \ref{fig:NMS3}.

\section{Experimental Validations}
\label{sec:ExperimentalValidations}

In this section, we first review the datasets used in our experiments.
Then, we present the implementation details and the evaluation method.
Finally, we report the results quantitatively and qualitatively.

\subsection{Datasets}
\label{sec:Dataset}

To evaluate the performance of the proposed approach for airplane detection in VHR remote sensing imagery, we use two publicly available datasets: NWPU VHR-10 \cite{cheng2014multi} and DOTA \cite{xia2018dota}.

NWPU VHR-10 is a collection of 800 very high resolution optical remote sensing images with 10 class objects that have been annotated in box-level.
These classes are airplane, ship, storage tank, baseball diamond, tennis court, basketball court, ground track field, harbor, bridge, and vehicle.
In this paper, we focus on the detection of airplanes that existed in 90 images with 757 instances.

DOTA is a large-scale dataset for object detection in aerial images.
It contains 1411 training and 458 validation aerial images from different sensors and platforms.
In this dataset, 15 objects are annotated by an oriented bounding box as well as a horizontal bounding box.
In this paper, we focus on the detection of airplanes that existed in 198 training images and 71 validation images, with 8071 and 2550 instances.
The physical size of one image pixel (i.e., gsd) in these 269 images varies from 0.09m to 4.2m (0.40m \(\pm\) 0.36m).

\begin{figure}[!t]
	\centering
	\subfigure[]
	{
		\includegraphics[width=.45\linewidth]{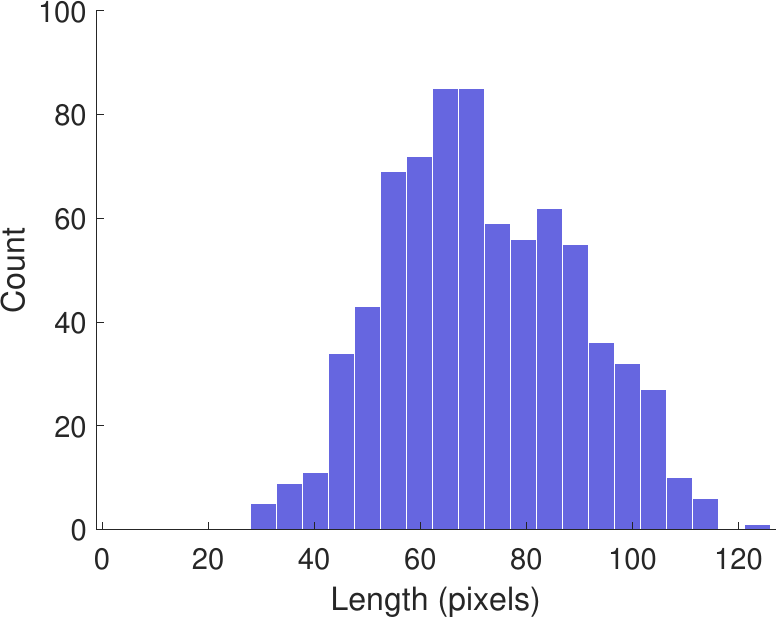}
		\label{fig:LengthHistogram_NWPU}
	}
	\subfigure[]
	{
		\includegraphics[width=.45\linewidth]{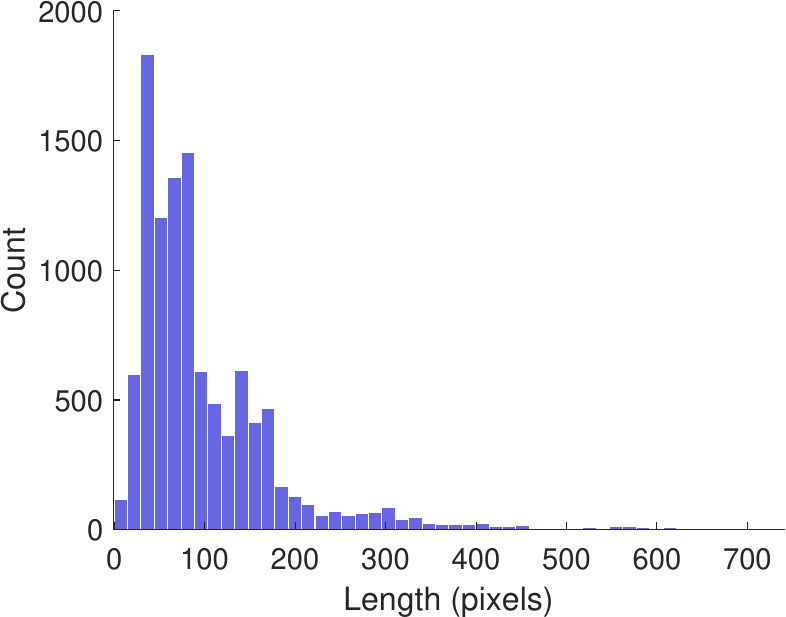}
		\label{fig:LengthHistogram_DOTA}
	}
	
	\caption{Histogram of the airplanes' length in (a) NWPU VHR-10 dataset, and (b) DOTA dataset}
	\label{fig:LengthHistogram}
\end{figure}

Two of the challenges in these datasets are the variety of scale and direction of the airplanes.
In aerial images, the direction of objects is not controllable, and so the algorithm should be rotation-invariant.
Moreover, the size of airplanes and the resolution of the imaging systems are different.
In NWPU VHR-10 dataset, 715 color images were acquired from Google Earth with the spatial resolution ranging from 0.5m to 2m, and 85 pansharpened color infrared images were acquired from Vaihingen data with a spatial resolution of 0.08m.
To show the variation in the length of the airplanes, we approximate the length of each airplane with the geometric mean of its box's width and height.
The histogram of approximated lengths for 757 airplanes is plotted in Fig. \ref{fig:LengthHistogram_NWPU}.
The minimum and maximum lengths are 30 and 126 pixels.
Similarly, we plot the histogram of lengths for 10621 airplanes of DOTA dataset in Fig. \ref{fig:LengthHistogram_DOTA}.
As can be observed, DOTA dataset is more challenging in which the lengths of airplanes varies from 7 to 740 pixels.

\begin{table*}\small
	\centering
	\caption{Detailed results of the proposed approach on NWPU VHR-10 dataset}
	\label{tab:ResultsStepsNWPU}
	\renewcommand{\arraystretch}{1.5}
	\begin{tabularx}{\linewidth}{|X|X|X|X|X|X|X|X|X|X|X|X|}
		\hline
		Fold	& Step		& Pos. 	& Neg. 		& Ind. 	& TP	& FN 	& FP 			& Rec. & Prec. & F1 & AP(\%)\\ \hlinewd{1.5pt}
			\multirow{3}{*}{Fold 1}
			& SLS		 	    &	5496	&	162786	& 2445	& 255	& 0		& 170472	& & &	& \\ \cline{2-12}
			& CNN					&	2838	&	19			&	17		& 255	&	0		& 2619		& & &	& \\ \cline{2-12}
			& NMS					&	255		&	2				&	0			& 255	&	0		& 2				& 1.000 & 0.992 & 0.996	& 99.99\\ \hlinewd{1.5pt}
			\multirow{3}{*}{Fold 2}
			& SLS		 	    & 5149	& 161509	& 2351	& 250	& 0		&	168759	& & &	& \\ \cline{2-12}
			& CNN					& 2607	& 69			& 15		& 249	&	1		&	2442		& & &	& \\ \cline{2-12}
			& NMS					&	249		&	6				& 0			& 249	&	1		&	6				& 0.996	& 0.976	& 0.986	& 99.46	\\ \hlinewd{1.5pt}
			\multirow{3}{*}{Fold 3}
			& SLS		 	    & 5689	& 162815	& 2393	& 252	& 0		&	170645	& & &	& \\ \cline{2-12}
			& CNN					& 2834	& 11			& 6			& 248	&	4		&	2603		& & &	& \\ \cline{2-12}
			& NMS					&	248		&	1				& 0			& 248	&	4		&	1				& 0.984	& 0.996	& 0.990	& 99.06	\\ \hlinewd{1.5pt}
			\multirow{3}{*}{Total}
			& SLS		 	    & 16334	& 487110	& 7839	& 757	& 0		&	509876	& & &	& \\ \cline{2-12}
			& CNN					& 8279	& 99			& 29		& 752	&	5		&	7664		& & & &	\\ \cline{2-12}
			& NMS					&	752		&	9				& 0			& 752	&	5		&	9				& 0.993	& 0.988	& 0.991	& 99.49	\\ \hlinewd{1.5pt}
	\end{tabularx}
	\begin{tablenotes}\small
		\item[*] * Pos.: Positive Instances, Neg.: Negative Instances, Ind.: Indeterminate Instances, Rec.: Recall, Prec.: Precision
  \end{tablenotes}
\end{table*}

\subsection{Implementation Details}
\label{sec:ImplementationDetails}

The proposed approach consists of three modules depicted in Fig. \ref{fig:ProposedApproach}.
The first module is the SLS proposal generator.
In SLS, we use the LSD algorithm for line segment detection, which works only on gray-scale images.
Hence, it is required to convert the RGB images to gray-scale ones, but what is the best RGB to gray conversion?
In RGB to gray conversion, \(2^{24}\) colors are mapped to only \(2^8\) numbers (on average, every \(2^{16}=65536\) colors are assigned to the same number).
Therefore, some of the boundaries visible in the RGB image may disappear in the new gray space.
For this reason, on NWPU VHR-10 dataset, we run the SLS proposal generator in each of the bands individually and then put all proposals together.
This will increase the number of proposals (roughly three times) and the recall value.
On the other hand, DOTA dataset is more challenging contains large images with very small to huge airplanes.
To achieve high recall and reduce computational complexity, we run the LSD algorithm with two sets of parameters only on the gray images.

The second module is the supervised classification of the proposals.
As mentioned, we use a deep convolutional neural network for feature extraction and classification.
In recent years, several different architectures are proposed to use in the classification tasks.
We employ the convolutional partition of ResNet50 architecture \cite{he2016deep} for feature extraction followed by a fully connected layer with and two neurons.
As is well known, 757 positive bags are insufficient for training such a network with \(23,459,010\) trainable parameters.
So, we use the convolutional layers of ResNet50 pre-trained on the ImageNet dataset \cite{deng2009imagenet} contained more than 14 million images.
Then, the pre-trained convolutional layers and the randomly-initialized fully connected layer are domain-specific fine-tuned for airplane detection.
For implementation, we use the Keras API \cite{chollet2015keras} and Adam optimizer \cite{kingma2014adam}.

Non-maximum suppression is the third module used only in the test phase.
In this module, we implement an iterative procedure in which we select the best proposal with the maximum score and remove the overlapped proposals with IoU greater than 0.25.

\subsection{Evaluation Method}
\label{sec:EvaluationMethod}

To evaluate the performance of the proposed approach on NWPU VHR-10 dataset, we use the 3-fold cross-validation method.
In this validation method, the dataset is split into three folds.
Then, three independent experiments are done in which two folds used for training of the model and the held-out used for validation.
However, since objects in an aerial image have similar resolution, brightness, angle of view, and geographic characteristics, we fold the dataset at the image level to achieve fair results.
In other words, each fold consists of all proposals generated from one-third of the images (i.e., 30 images).
This allows us to investigate how well our system works on new, unseen images.
In DOTA dataset, the training and the validation images are specified, so we run only one experiment.

We use \(Recall\), \(Precision\), and \(F_1\) score to evaluate the trained model with the following equations:

\begin{align}
	& Recall = \frac{TP}{TP + FN} \\
	& Precision = \frac{TP}{TP + FP} \\
	& F_1 = \frac{2\times Recall\times Precision}{Recall + Precision}
\end{align}

where TP (true positive) and FN (false negative) are the numbers of detected and missed airplanes, respectively.
Also, FP (false positive) is the number of other objects that are mistakenly identified as an airplane.
\(Recall\) is the fraction of relevant instances that have been detected over the total amount of airplanes, while \(Precision\) is the fraction of relevant instances among the detected instances.
Both \(Recall\) and \(Precision\) take only one side of the algorithm, while \(F_1\) considers both of them to compute the overall score.

Another evaluation metric widely used in the object detection domain is the average precision (\(AP\)).
The \(AP\) computes the average value of the \(Precision\) over the interval from \(Recall\) = 0 to
\(Recall\) = 1 (i.e., the area under the \(Precision-Recall\) curve).

\begin{figure*}[t]
	\hspace*{.1in}
	\includegraphics[width=.8\textwidth,left]{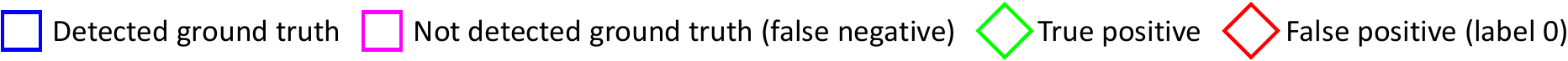}
	\centering
	\subfigure
	{
		\includegraphics[width=0.48\linewidth]{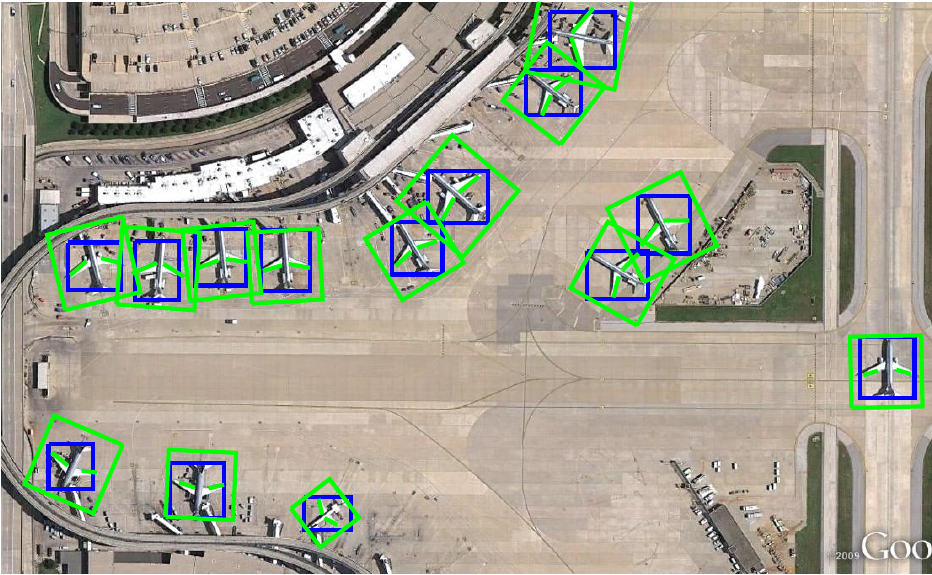} 
		\includegraphics[width=0.48\linewidth]{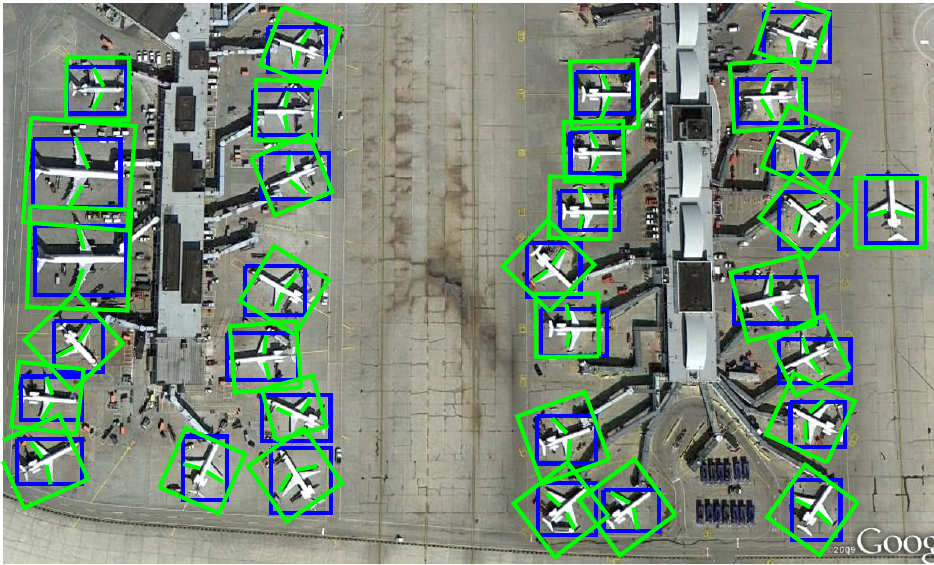} 
	}
	\subfigure
	{
		\includegraphics[width=0.49\linewidth]{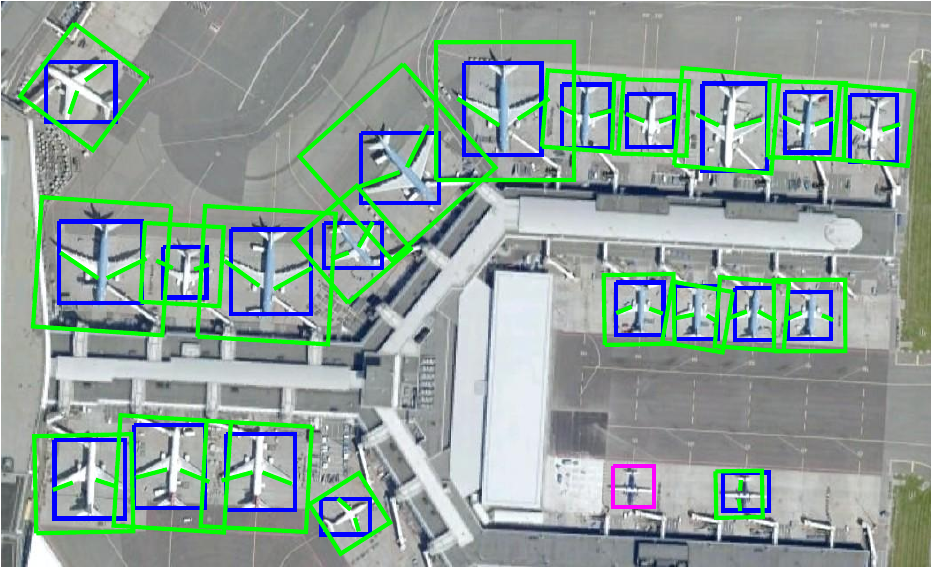} 
		\includegraphics[width=0.49\linewidth]{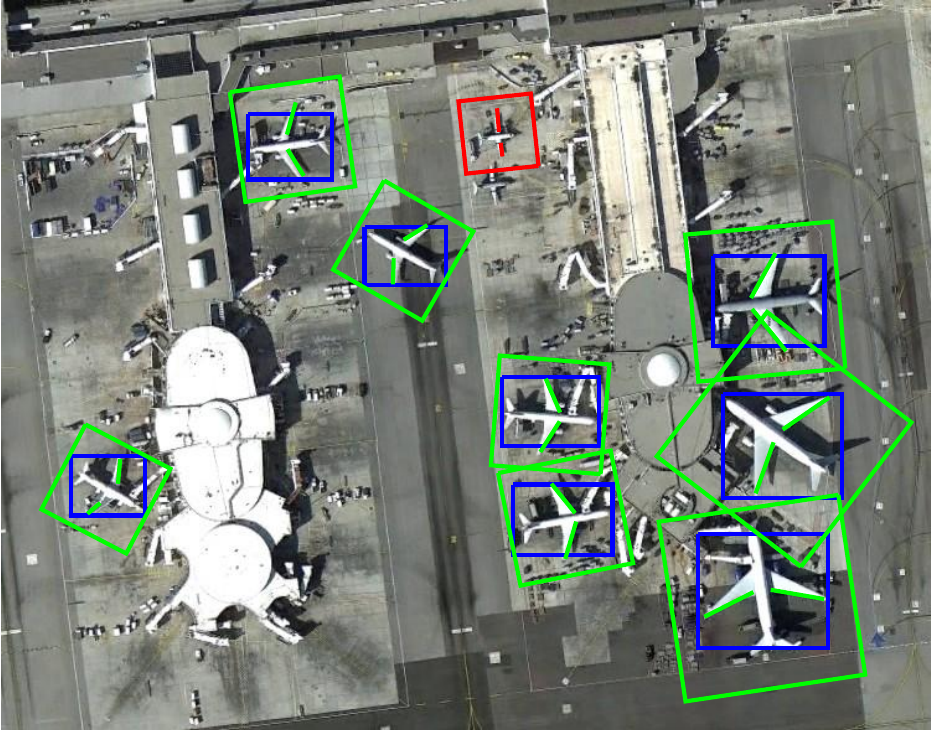}
	}
	\caption{Sample results of the proposed airplane detection algorithm on NWPU VHR-10 dataset.}
	\label{fig:Results}
\end{figure*}

\subsection{Performance Evaluation}
\label{sec:PerformanceEvaluation}

The detailed results of the modules of the proposed approach on NWPU VHR-10 dataset are given in Table \ref{tab:ResultsStepsNWPU}.
In this table, positive, negative and indeterminate instances are the number of instances with labels \(+1\), \(0\), and \(-1\), respectively.
TP is the number of ground truth airplanes with at least one corresponding positive instance.
FN is the number of ground truth airplanes with no corresponding positive instance.
FP is the number of all instances minus TP (i.e., from the corresponding positive instances with one ground truth, only one is considered as true positive and the others are counted as false positive).

As can be observed in Table \ref{tab:ResultsStepsNWPU}, for all airplanes with different scales and directions, at least one appropriate candidate is extracted by the SLS proposal generator (on average, about 21 positive instances for each airplane).
These positive instances are formed from different symmetric line segments in an airplane in three bands.
Some positive instances are correctly aligned by the airplane's fuselage while others are not in that direction (this challenge is mainly due to the box-level annotation of the ground truth).
Because of this, we used the MIL idea for the training of the network so that the network is allowed to learn at-least one positive instance from all positive instances corresponding to one airplane.
By comparing the first and the second rows in each fold, we find that the network rejects about \(50\%\) of the positive instances while misses about \(1\%\) of the positive bags (i.e., airplanes).

Non-maximum suppression (NMS) is the last module of the proposed approach, which is very important due to the multiple detections of each airplane.
From the results reported in Table \ref{tab:ResultsStepsNWPU}, we find the the NMS step has worked well on NWPU VHR-10 dataset.

Altogether, more than \(99\%\) of the airplanes are detected by the proposed approach while only \(9\) false alarms are produced (i.e., \(Precision = 0.988\)).
Some detection results of the proposed approach are shown in Fig. \ref{fig:Results}.
As can be observed, in addition to the high performance of the proposed algorithm for detecting airplanes, for most of them, the direction is correctly estimated.
As a result, one of the main advantages of the proposed algorithm is the ability to estimate airplane direction using the box-level annotations.

\begin{table*}\small
	\centering
	\caption{Detailed results of the proposed approach on DOTA dataset}
	\label{tab:ResultsStepsDOTA}
	\renewcommand{\arraystretch}{1.5}
	\begin{tabularx}{\linewidth}{|X|X|X|X|X|X|X|X|X|X|X|}
		\hline
			Step				& Pos. 	& Neg. 		& Ind. 	& TP		& FN 	& FP 			& Rec. & Prec. & F1 & AP (\%)\\ \hlinewd{1.5pt}
			SLS		 	    &	37647	&	4518906	& 21049	& 2475 	& 75	& 4575127	& & &	& \\ \hline
			CNN					&	15341	&	749			&	167		& 2393	&	157	& 13864		& & &	& \\ \hline
			NMS					&	2368	&	235			&	8			& 2363	&	187	& 248			& 0.918 & 0.925 & 0.922	& 92.53\\ \hlinewd{1.5pt}
		\end{tabularx}
	\begin{tablenotes}\small
		\item[*] * Pos.: Positive Instances, Neg.: Negative Instances, Ind.: Indeterminate Instances, Rec.: Recall, Prec.: Precision
  \end{tablenotes}
\end{table*}

\begin{figure*}[htb]
	\hspace*{.2in}
	\includegraphics[width=.8\textwidth,left]{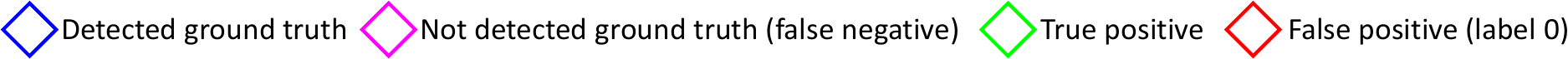}
	\centering
	\subfigure
	{
		\includegraphics[width=0.150\linewidth]{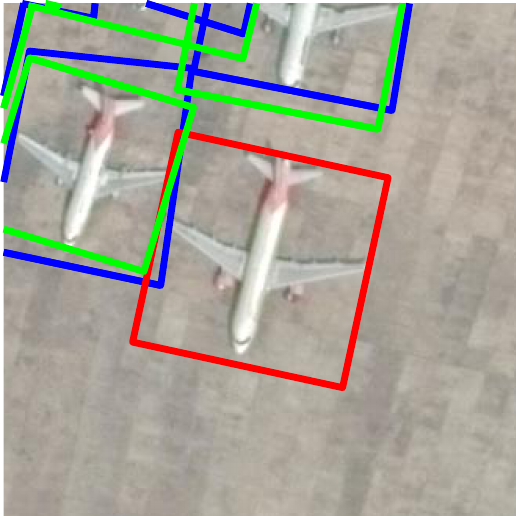}
		\includegraphics[width=0.150\linewidth]{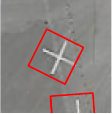}
		\includegraphics[width=0.150\linewidth]{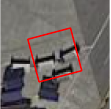}
		\includegraphics[width=0.150\linewidth]{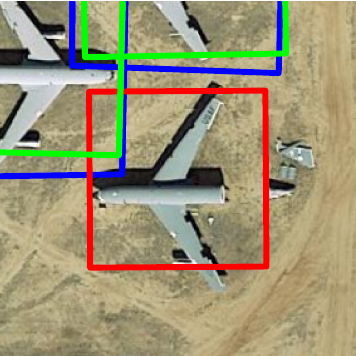}
		\includegraphics[width=0.150\linewidth]{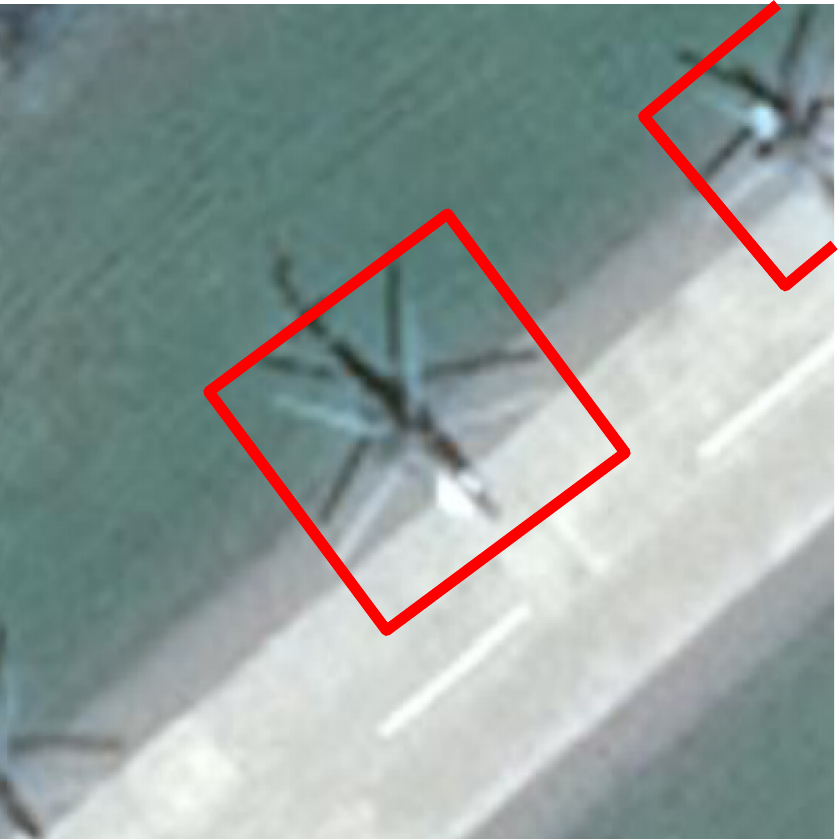}
		\includegraphics[width=0.150\linewidth]{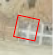}
	}
	\subfigure
	{
		\includegraphics[width=0.150\linewidth]{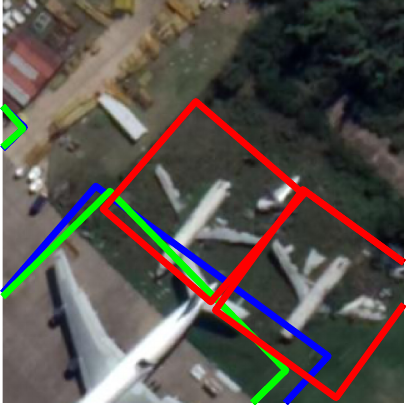}
		\includegraphics[width=0.150\linewidth]{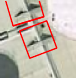}
		\includegraphics[width=0.150\linewidth]{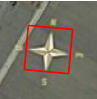}
		\includegraphics[width=0.150\linewidth]{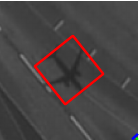}
		\includegraphics[width=0.150\linewidth]{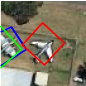}
		\includegraphics[width=0.150\linewidth]{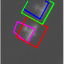}
	}
	\subfigure
	{
		\includegraphics[width=0.150\linewidth]{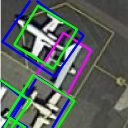}
		\includegraphics[width=0.150\linewidth]{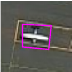}
		\includegraphics[width=0.150\linewidth]{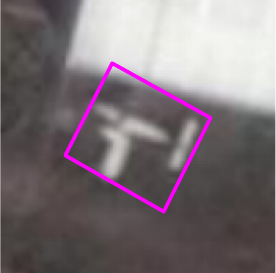}
		\includegraphics[width=0.150\linewidth]{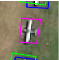}
		\includegraphics[width=0.150\linewidth]{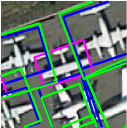}
		\includegraphics[width=0.150\linewidth]{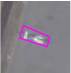}
	}
	\subfigure
	{
		\includegraphics[width=0.150\linewidth]{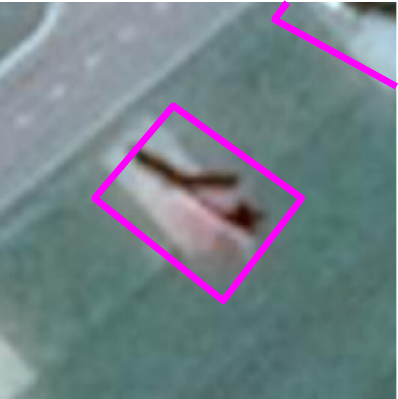}
		\includegraphics[width=0.150\linewidth]{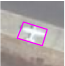}
		\includegraphics[width=0.150\linewidth]{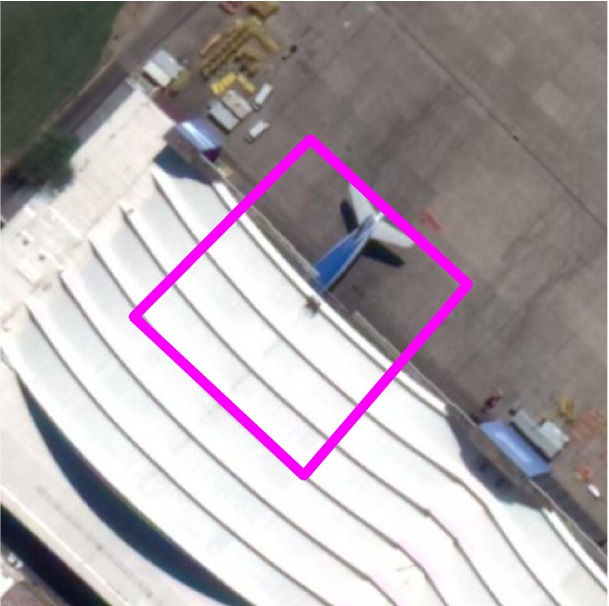}
		\includegraphics[width=0.150\linewidth]{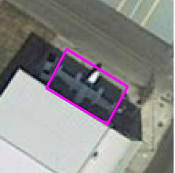}
		\includegraphics[width=0.150\linewidth]{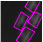}
		\includegraphics[width=0.150\linewidth]{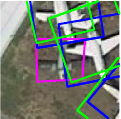}
	}
	\caption{Sample errors of the proposed airplane detection algorithm on DOTA dataset.}
	\label{fig:ErrorsDOTA}
\end{figure*}

We have reported the detailed results of the proposed approach on DOTA dataset (validation subset) in Table \ref{tab:ResultsStepsDOTA}.
Due to the high dimensions of most images in this dataset, the number of proposals is much more than NWPU dataset.
Moreover, the high density of airplanes in some areas has caused some of them missed in the NMS step.
We demonstrate some errors of the proposed algorithm on DOTA dataset in Fig. \ref{fig:ErrorsDOTA}.
As can be seen, the errors are very challenging samples, including some wrong labeling in the dataset.

\subsubsection{Computational Cost}
\label{sec:ComputationalCost}

We ran the experiments on a computer with GeForce GTX1050 GPU, Intel Core i7 CPU, and 16GB RAM
The average running time per image of the proposed algorithm on NWPU VHR-10 dataset is reported in Table \ref{tab:Time}.
In this table, we present the running time of each module in two proposal generation setups: using only the gray image in comparison to using three bands (RGB).
As expected, the number of proposals, and consequently, the processing time in the RGB setup, is approximately three times the gray setup.
On the other hand, the number of missed airplanes is 8 and 5, in the RGB and gray, respectively, with the same false alarms.
So, we can use the algorithm with \(AP = 99.29\%\) at 0.19 fps or \(AP=99.49\%\) at 0.06 fps.

\begin{table}\small
	\centering
	\caption{Computational Cost of the proposed algorithm on NWPU VHR-10 dataset}
	\label{tab:Time}
	\renewcommand{\arraystretch}{1.5}
	\begin{tabularx}{\linewidth}{|X|X|X|X|}
		\hline
			& 	& RGB 	& Gray \\ \hlinewd{1.5pt}
			\multirow{4}{*}{\shortstack{Average \\ Time (sec)}}
			& SLS		 	    &	7.76	& 2.54	\\ \cline{2-4}
			& CNN					&	8.02	&	2.63	\\ \cline{2-4}
			& NMS					&	0.02	&	0.01	\\ \cline{2-4}
			& All					&	15.80	&	5.18	\\ \hlinewd{1.5pt}
			\multirow{4}{*}{Performance}
			& Recall		 	&	0.993	&	0.989	\\ \cline{2-4}
			& Precision		&	0.988	&	0.988	\\ \cline{2-4}
			& F1					&	0.991	&	0.989	\\ \cline{2-4}
			& AP (\%)			& 99.49	&	99.29 \\ \hlinewd{1.5pt}
	\end{tabularx}
\end{table}

\subsubsection{Comparison with Previous Works}
\label{sec:ComparisonWithThePreviousWorks}

Unfortunately, the experimental setups of the airplane detection algorithms in the literature are very different, which makes it difficult to compare them fairly.
For example, the datasets used in \cite{yu2015rotation,yu2016rotation,long2017accurate} are collected by the authors and are not publicly available.
Also, different evaluation measures such as \(AP\) and \(F_{1}\) have reported.
Besides, various data partitioning methods, such as k-fold and holdout, have used.

Considering these differences, the reported results of some previous works on NWPU VHR-10 dataset, as well as our results, are presented in Table 1.
According to this table, our results are promising and are among the best ones for airplane detection.
Notably, Xu et al. in \cite{xu2020hierarchical} used the holdout method (i.e., 20\% for training, 20\% for validation, and 60\% for testing) while we used 3-fold cross-validation.
So, the selected images for validation in the holdout method are crucial, especially considering this fact that all errors of our algorithm occurred in only 11 images from 90.

Using box-level annotations, most of the existing studies have focused on detecting the bounding box of airplanes.
In contrast, as a significant advantage, our algorithm can estimate the direction of airplanes.

\begin{table}\small
	\centering
	\caption{Comparison with previous works on NWPU VHR-10 dataset}
	\label{tab:Comparison}
	\renewcommand{\arraystretch}{1.5}
		\begin{tabularx}{\linewidth}{|X|X|X|} 
		\hline
		Method												& F1			& AP (\%) 		\\ \hlinewd{1.5pt}
			\cite{cheng2016object}			& -				& 66.31	\\ \hline
			\cite{xu2017deformable}			& -				& 87.3	\\ \hline
			\cite{cheng2016learning}		& -				& 88.35	\\ \hline
			\cite{zhong2018multi}				& -				& 90.7	\\ \hline
			\cite{wang2018two}					& -				& 90.91	\\ \hline
			\cite{zou2018random}				& -				& 94.1	\\ \hline
			\cite{zhang2019multi}				& -				& 95.79	\\ \hline
			\cite{deng2018multi}				& -				& 98.73	\\ \hline
			\cite{chen2019multi}				& -				& 99.47	\\ \hline
			\cite{xu2020hierarchical}		& -				& \texttt{99.79}	\\ \hline
			\cite{qiu2017occluded}			& 0.912		& -			\\ \hline
			\cite{qiu2017automatic}			& 0.917		& -			\\ \hline
			\cite{qiu2018unified}				& 0.920		& -			\\ \hline
			Ours												& \textbf{0.992}		& 99.49	\\ \hline
	\end{tabularx}
\end{table}


DOTA is a large-scale dataset with specified training and validation subsets, and consequently, the comparison is more fair.
We report the results of some recent works on DOTA dataset in comparison to our results in Table \ref{tab:ComparisonDOTA}.
From this table, it can be observed that the proposed algorithm is promising for airplane detection in challenging situations.

\section{Conclusion}
\label{sec:Conclusion}

In this paper, we have presented a rotation-and-scale invariant airplane detection algorithm that consists of three main modules: proposal generation, deep classification, and non-maximum suppression.

For proposal generation, we introduced a new method based on the symmetric line segments.
In this method, we have employed the common properties of the boundaries of the airplanes from the top view that can be approximated by a chain of symmetric line segments.
As have reported in the experimental results, this method is promising for airplane candidate extraction with arbitrary direction and in a wide range of scales.

In NWPU VHR-10 dataset, the airplanes are annotated in the box-level, which is much simpler and more practical than pixel-level annotations.
Therefore, the separation of the symmetric line segments that are correctly in the direction of the airplane is not possible from other overlapping proposals with a ground-truth box.
So, in the second module, we have used a multiple instance learning based loss function to train the deep convolutional neural network.
More precisely, we allow the network to learn at least one proposal from the proposals corresponds to one ground-truth airplane.
We have implemented this idea successfully, and as a significant result, we can estimate the direction of most of the airplanes correctly without being specified in the dataset.

Since each airplane may have more than one appropriate symmetric line segment, as the third module in the test phase, 
we have used the non-maximum suppression algorithm.
Experiments conducted on NWPU VHR-10 and DOTA datasets have shown that the proposed algorithm obtained \(99.49\%\) and \(92.53\%\) average precision, respectively.

As can be observed in Fig. \ref{fig:ErrorsDOTA}, 
most false negatives of the proposed algorithm are airplanes whose symmetrical line segments are not correctly detected.
So, our future work is to design a network for proposal generation based on the symmetric line segments idea, using the recent deep learning-based line detection algorithms.
So, we will be able to train the full network (proposal generation and classification) in an end-to-end manner.

\begin{table}\small
	\centering
	\caption{Comparison with previous works on DOTA dataset}
	\label{tab:ComparisonDOTA}
	\renewcommand{\arraystretch}{1.5}
		\begin{tabularx}{\linewidth}{|X|X|} 
		\hline
		Method												& AP 		\\ \hlinewd{1.5pt}
			\cite{ding2018learning}			& 88.64	\\ \hline
			\cite{zhang2019geospatial}	& 90.07	\\ \hline
			\cite{xu2020hierarchical}		& 90.39	\\ \hline
			Ours												& \texttt{92.53}	\\ \hline
	\end{tabularx}
\end{table}

\bibliographystyle{elsarticle-num}

\bibliography{References}

\begin{thebibliography}{10}
\expandafter\ifx\csname url\endcsname\relax
  \def\url#1{\texttt{#1}}\fi
\expandafter\ifx\csname urlprefix\endcsname\relax\def\urlprefix{URL }\fi
\expandafter\ifx\csname href\endcsname\relax
  \def\href#1#2{#2} \def\path#1{#1}\fi

\bibitem{seelan2003remote}
S.~K. Seelan, S.~Laguette, G.~M. Casady, G.~A. Seielstad, Remote sensing
  applications for precision agriculture: A learning community approach, Remote
  Sensing of Environment 88~(1-2) (2003) 157--169.

\bibitem{james2019below}
J.~James, J.~J. Ford, T.~L. Molloy, Below horizon aircraft detection using deep
  learning for vision-based sense and avoid, in: 2019 International Conference
  on Unmanned Aircraft Systems (ICUAS), IEEE, 2019, pp. 965--970.

\bibitem{khan2018modern}
M.~J. Khan, H.~S. Khan, A.~Yousaf, K.~Khurshid, A.~Abbas, Modern trends in
  hyperspectral image analysis: a review, IEEE Access 6 (2018) 14118--14129.

\bibitem{li2020deep}
Z.~Li, K.~Xu, J.~Xie, Q.~Bi, K.~Qin, Deep multiple instance convolutional
  neural networks for learning robust scene representations, IEEE Transactions
  on Geoscience and Remote Sensing 58~(5) (2020) 3685--3702.

\bibitem{sizkouhi2020automatic}
A.~M.~M. Sizkouhi, M.~Aghaei, S.~M. Esmailifar, M.~R. Mohammadi, F.~Grimaccia,
  Automatic boundary extraction of large-scale photovoltaic plants using a
  fully convolutional network on aerial imagery, IEEE Journal of Photovoltaics
  10~(4) (2020) 1061--1067.

\bibitem{cheng2016survey}
G.~Cheng, J.~Han, A survey on object detection in optical remote sensing
  images, ISPRS Journal of Photogrammetry and Remote Sensing 117 (2016) 11--28.

\bibitem{krizhevsky2012imagenet}
A.~Krizhevsky, I.~Sutskever, G.~E. Hinton, Imagenet classification with deep
  convolutional neural networks, in: Advances in neural information processing
  systems, 2012, pp. 1097--1105.

\bibitem{simonyan2014very}
K.~Simonyan, A.~Zisserman, Very deep convolutional networks for large-scale
  image recognition, arXiv preprint arXiv:1409.1556 (2014).

\bibitem{huang2017densely}
G.~Huang, Z.~Liu, L.~Van Der~Maaten, K.~Q. Weinberger, Densely connected
  convolutional networks, in: 2017 IEEE Conference on Computer Vision and
  Pattern Recognition (CVPR), 2017, pp. 2261--2269.
\newblock \href {https://doi.org/10.1109/CVPR.2017.243}
  {\path{doi:10.1109/CVPR.2017.243}}.

\bibitem{amores2013multiple}
J.~Amores, Multiple instance classification: Review, taxonomy and comparative
  study, Artificial Intelligence 201 (2013) 81--105.

\bibitem{cheng2014multi}
G.~Cheng, J.~Han, P.~Zhou, L.~Guo, Multi-class geospatial object detection and
  geographic image classification based on collection of part detectors, ISPRS
  Journal of Photogrammetry and Remote Sensing 98 (2014) 119--132.

\bibitem{xia2018dota}
G.-S. Xia, X.~Bai, J.~Ding, Z.~Zhu, S.~Belongie, J.~Luo, M.~Datcu, M.~Pelillo,
  L.~Zhang, Dota: A large-scale dataset for object detection in aerial images,
  in: Proceedings of the IEEE Conference on Computer Vision and Pattern
  Recognition, 2018, pp. 3974--3983.

\bibitem{sun2012automatic}
H.~Sun, X.~Sun, H.~Wang, Y.~Li, X.~Li, Automatic target detection in
  high-resolution remote sensing images using spatial sparse coding
  bag-of-words model, IEEE Geoscience and Remote Sensing Letters 9~(1) (2012)
  109--113.

\bibitem{redmon2016you}
J.~Redmon, S.~Divvala, R.~Girshick, A.~Farhadi, You only look once: Unified,
  real-time object detection, in: Proceedings of the IEEE conference on
  computer vision and pattern recognition, 2016, pp. 779--788.

\bibitem{liu2016ssd}
W.~Liu, D.~Anguelov, D.~Erhan, C.~Szegedy, S.~Reed, C.-Y. Fu, A.~C. Berg, Ssd:
  Single shot multibox detector, in: European conference on computer vision,
  Springer, 2016, pp. 21--37.

\bibitem{bo2010region}
S.~Bo, Y.~Jing, Region-based airplane detection in remotely sensed imagery, in:
  Image and Signal Processing (CISP), 2010 3rd International Congress on,
  Vol.~4, IEEE, 2010, pp. 1923--1926.

\bibitem{li2012automatic}
Y.~Li, X.~Sun, H.~Wang, H.~Sun, X.~Li, Automatic target detection in
  high-resolution remote sensing images using a contour-based spatial model,
  IEEE Geoscience and Remote Sensing Letters 9~(5) (2012) 886--890.

\bibitem{zitnick2014edge}
C.~L. Zitnick, P.~Doll{\'a}r, Edge boxes: Locating object proposals from edges,
  in: European conference on computer vision, Springer, 2014, pp. 391--405.

\bibitem{uijlings2013selective}
J.~R. Uijlings, K.~E. Van De~Sande, T.~Gevers, A.~W. Smeulders, Selective
  search for object recognition, International journal of computer vision
  104~(2) (2013) 154--171.

\bibitem{cheng2014bing}
M.-M. Cheng, Z.~Zhang, W.-Y. Lin, P.~Torr, Bing: Binarized normed gradients for
  objectness estimation at 300fps, in: Proceedings of the IEEE conference on
  computer vision and pattern recognition, 2014, pp. 3286--3293.

\bibitem{humayun2014rigor}
A.~Humayun, F.~Li, J.~M. Rehg, Rigor: Reusing inference in graph cuts for
  generating object regions, in: Proceedings of the IEEE Conference on Computer
  Vision and Pattern Recognition, 2014, pp. 336--343.

\bibitem{hosang2016makes}
J.~Hosang, R.~Benenson, P.~Doll{\'a}r, B.~Schiele, What makes for effective
  detection proposals?, IEEE transactions on pattern analysis and machine
  intelligence 38~(4) (2016) 814--830.

\bibitem{khan2017automatic}
M.~J. Khan, A.~Yousaf, N.~Javed, S.~Nadeem, K.~Khurshid, Automatic target
  detection in satellite images using deep learning, Journal of Space
  Technology 7~(1) (2017) 44--49.

\bibitem{farooq2017efficient}
A.~Farooq, J.~Hu, X.~Jia, Efficient object proposals extraction for target
  detection in vhr remote sensing images, in: 2017 IEEE International
  Geoscience and Remote Sensing Symposium (IGARSS), IEEE, 2017, pp. 3337--3340.

\bibitem{li2018aircraft}
Y.~Li, K.~Fu, H.~Sun, X.~Sun, An aircraft detection framework based on
  reinforcement learning and convolutional neural networks in remote sensing
  images, Remote Sensing 10~(2) (2018) 243.

\bibitem{wu2015fast}
H.~Wu, H.~Zhang, J.~Zhang, F.~Xu, Fast aircraft detection in satellite images
  based on convolutional neural networks, in: Image Processing (ICIP), 2015
  IEEE International Conference on, IEEE, 2015, pp. 4210--4214.

\bibitem{luo2016airplane}
Q.~Luo, Z.~Shi, Airplane detection in remote sensing images based on object
  proposal, in: 2016 IEEE International Geoscience and Remote Sensing Symposium
  (IGARSS), IEEE, 2016, pp. 1388--1391.

\bibitem{zheng2016object}
J.~Zheng, Y.~Xi, M.~Feng, X.~Li, N.~Li, Object detection based on bing in
  optical remote sensing images, in: 2016 9th International Congress on Image
  and Signal Processing, BioMedical Engineering and Informatics (CISP-BMEI),
  IEEE, 2016, pp. 504--509.

\bibitem{long2017accurate}
Y.~Long, Y.~Gong, Z.~Xiao, Q.~Liu, Accurate object localization in remote
  sensing images based on convolutional neural networks, IEEE Transactions on
  Geoscience and Remote Sensing 55~(5) (2017) 2486--2498.

\bibitem{liu2020aircraft}
Q.~Liu, X.~Xiang, Y.~Wang, Z.~Luo, F.~Fang, Aircraft detection in remote
  sensing image based on corner clustering and deep learning, Engineering
  Applications of Artificial Intelligence 87 (2020) 103333.

\bibitem{kawato2001circle}
S.~Kawato, N.~Tetsutani, Circle-frequency filter and its application, in: ITE
  Technical Report 25.12, The Institute of Image Information and Television
  Engineers, 2001, pp. 49--54.

\bibitem{cai2006airplane}
H.~Cai, Y.~Su, Airplane detection in remote sensing image with a
  circle-frequency filter, in: International Conference on Space Information
  Technology, Vol. 5985, International Society for Optics and Photonics, 2006,
  p. 59852T.

\bibitem{gao2013aircraft}
F.~Gao, Q.~Xu, B.~Li, Aircraft detection from vhr images based on
  circle-frequency filter and multilevel features, The Scientific World Journal
  2013 (2013).

\bibitem{an2014automated}
Z.~An, Z.~Shi, X.~Teng, X.~Yu, W.~Tang, An automated airplane detection system
  for large panchromatic image with high spatial resolution,
  Optik-International Journal for Light and Electron Optics 125~(12) (2014)
  2768--2775.

\bibitem{zhang2015unsupervised}
W.~Zhang, W.~Lv, Y.~Zhang, J.~Tian, J.~Ma, Unsupervised-learning airplane
  detection in remote sensing images, in: MIPPR 2015: Remote Sensing Image
  Processing, Geographic Information Systems, and Other Applications, Vol.
  9815, International Society for Optics and Photonics, 2015, p. 981503.

\bibitem{hosomura2010airplane}
T.~Hosomura, et~al., Airplane extraction from high resolution satellite image
  using boundary feature, in: Proc. of ISPRS Technical Com. VIII Symposium,
  2010.

\bibitem{inglada2007automatic}
J.~Inglada, Automatic recognition of man-made objects in high resolution
  optical remote sensing images by svm classification of geometric image
  features, ISPRS journal of photogrammetry and remote sensing 62~(3) (2007)
  236--248.

\bibitem{dalal2005histograms}
N.~Dalal, B.~Triggs, Histograms of oriented gradients for human detection, in:
  Computer Vision and Pattern Recognition, 2005. CVPR 2005. IEEE Computer
  Society Conference on, Vol.~1, IEEE, 2005, pp. 886--893.

\bibitem{cheng2013object}
G.~Cheng, J.~Han, L.~Guo, X.~Qian, P.~Zhou, X.~Yao, X.~Hu, Object detection in
  remote sensing imagery using a discriminatively trained mixture model, ISPRS
  Journal of Photogrammetry and Remote Sensing 85 (2013) 32--43.

\bibitem{cheng2016object}
G.~Cheng, P.~Zhou, X.~Yao, C.~Yao, Y.~Zhang, J.~Han, Object detection in vhr
  optical remote sensing images via learning rotation-invariant hog feature,
  in: Earth Observation and Remote Sensing Applications (EORSA), 2016 4th
  International Workshop on, IEEE, 2016, pp. 433--436.

\bibitem{bai2014vhr}
X.~Bai, H.~Zhang, J.~Zhou, Vhr object detection based on structural feature
  extraction and query expansion, IEEE Transactions on Geoscience and Remote
  Sensing 52~(10) (2014) 6508--6520.

\bibitem{kumar2014detailed}
G.~Kumar, P.~K. Bhatia, A detailed review of feature extraction in image
  processing systems, in: Advanced Computing \& Communication Technologies
  (ACCT), 2014 Fourth International Conference on, IEEE, 2014, pp. 5--12.

\bibitem{lecun1998gradient}
Y.~LeCun, L.~Bottou, Y.~Bengio, P.~Haffner, Gradient-based learning applied to
  document recognition, Proceedings of the IEEE 86~(11) (1998) 2278--2324.

\bibitem{szegedy2015going}
C.~Szegedy, W.~Liu, Y.~Jia, P.~Sermanet, S.~Reed, D.~Anguelov, D.~Erhan,
  V.~Vanhoucke, A.~Rabinovich, Going deeper with convolutions, in: Proceedings
  of the IEEE conference on computer vision and pattern recognition, 2015, pp.
  1--9.

\bibitem{howard2017mobilenets}
A.~G. Howard, M.~Zhu, B.~Chen, D.~Kalenichenko, W.~Wang, T.~Weyand,
  M.~Andreetto, H.~Adam, Mobilenets: Efficient convolutional neural networks
  for mobile vision applications, arXiv preprint arXiv:1704.04861 (2017).

\bibitem{tan2019efficientnet}
M.~Tan, Q.~V. Le, Efficientnet: Rethinking model scaling for convolutional
  neural networks, arXiv preprint arXiv:1905.11946 (2019).

\bibitem{zhong2018multi}
Y.~Zhong, X.~Han, L.~Zhang, Multi-class geospatial object detection based on a
  position-sensitive balancing framework for high spatial resolution remote
  sensing imagery, ISPRS Journal of Photogrammetry and Remote Sensing 138
  (2018) 281--294.

\bibitem{deng2018multi}
Z.~Deng, H.~Sun, S.~Zhou, J.~Zhao, L.~Lei, H.~Zou, Multi-scale object detection
  in remote sensing imagery with convolutional neural networks, ISPRS Journal
  of Photogrammetry and Remote Sensing (2018).

\bibitem{zou2018random}
Z.~Zou, Z.~Shi, Random access memories: A new paradigm for target detection in
  high resolution aerial remote sensing images, IEEE Transactions on Image
  Processing 27~(3) (2018) 1100--1111.

\bibitem{li2011saliency}
Z.~Li, L.~Itti, Saliency and gist features for target detection in satellite
  images, IEEE Transactions on Image Processing 20~(7) (2011) 2017--2029.

\bibitem{cheng2014scalable}
G.~Cheng, J.~Han, P.~Zhou, L.~Guo, Scalable multi-class geospatial object
  detection in high-spatial-resolution remote sensing images, in: Geoscience
  and Remote Sensing Symposium (IGARSS), 2014 IEEE International, IEEE, 2014,
  pp. 2479--2482.

\bibitem{cheng2016learning}
G.~Cheng, P.~Zhou, J.~Han, Learning rotation-invariant convolutional neural
  networks for object detection in vhr optical remote sensing images, IEEE
  Transactions on Geoscience and Remote Sensing 54~(12) (2016) 7405--7415.

\bibitem{petridis2008learning}
S.~Petridis, C.~Geyer, S.~Singh, Learning to detect aircraft at low
  resolutions, in: International Conference on Computer Vision Systems,
  Springer, 2008, pp. 474--483.

\bibitem{duda1972use}
R.~O. Duda, P.~E. Hart, Use of the hough transformation to detect lines and
  curves in pictures, Communications of the ACM 15~(1) (1972) 11--15.

\bibitem{von2010lsd}
R.~G. Von~Gioi, J.~Jakubowicz, J.-M. Morel, G.~Randall, Lsd: A fast line
  segment detector with a false detection control, IEEE transactions on pattern
  analysis and machine intelligence 32~(4) (2010) 722--732.

\bibitem{lin2018focal}
T.-Y. Lin, P.~Goyal, R.~Girshick, K.~He, P.~Doll{\'a}r, Focal loss for dense
  object detection, IEEE Transactions on Pattern Analysis and Machine
  Intelligence (2018).
\newblock \href {https://doi.org/10.1109/TPAMI.2018.2858826}
  {\path{doi:10.1109/TPAMI.2018.2858826}}.

\bibitem{he2016deep}
K.~He, X.~Zhang, S.~Ren, J.~Sun, Deep residual learning for image recognition,
  in: Proceedings of the IEEE conference on computer vision and pattern
  recognition, 2016, pp. 770--778.

\bibitem{deng2009imagenet}
J.~Deng, W.~Dong, R.~Socher, L.-J. Li, K.~Li, L.~Fei-Fei, Imagenet: A
  large-scale hierarchical image database, in: Computer Vision and Pattern
  Recognition, 2009. CVPR 2009. IEEE Conference on, Ieee, 2009, pp. 248--255.

\bibitem{chollet2015keras}
F.~Chollet, et~al., Keras, \url{https://keras.io} (2015).

\bibitem{kingma2014adam}
D.~P. Kingma, J.~Ba, Adam: A method for stochastic optimization, arXiv preprint
  arXiv:1412.6980 (2014).

\bibitem{yu2015rotation}
Y.~Yu, H.~Guan, Z.~Ji, Rotation-invariant object detection in high-resolution
  satellite imagery using superpixel-based deep hough forests, IEEE Geoscience
  and Remote Sensing Letters 12~(11) (2015) 2183--2187.

\bibitem{yu2016rotation}
Y.~Yu, H.~Guan, D.~Zai, Z.~Ji, Rotation-and-scale-invariant airplane detection
  in high-resolution satellite images based on deep-hough-forests, ISPRS
  Journal of Photogrammetry and Remote Sensing 112 (2016) 50--64.

\bibitem{xu2020hierarchical}
C.~Xu, C.~Li, Z.~Cui, T.~Zhang, J.~Yang, Hierarchical semantic propagation for
  object detection in remote sensing imagery, IEEE Transactions on Geoscience
  and Remote Sensing (2020).

\bibitem{xu2017deformable}
Z.~Xu, X.~Xu, L.~Wang, R.~Yang, F.~Pu, Deformable convnet with aspect ratio
  constrained nms for object detection in remote sensing imagery, Remote
  Sensing 9~(12) (2017) 1312.

\bibitem{wang2018two}
S.~Wang, M.~Wang, X.~Zhao, D.~Liu, Two-stage object detection based on deep
  pruning for remote sensing image, in: International Conference on Knowledge
  Science, Engineering and Management, Springer, 2018, pp. 137--147.

\bibitem{zhang2019multi}
W.~Zhang, L.~Jiao, X.~Liu, J.~Liu, Multi-scale feature fusion network for
  object detection in vhr optical remote sensing images, in: IGARSS 2019-2019
  IEEE International Geoscience and Remote Sensing Symposium, IEEE, 2019, pp.
  330--333.

\bibitem{chen2019multi}
J.~Chen, L.~Wan, J.~Zhu, G.~Xu, M.~Deng, Multi-scale spatial and channel-wise
  attention for improving object detection in remote sensing imagery, IEEE
  Geoscience and Remote Sensing Letters (2019).

\bibitem{qiu2017occluded}
S.~Qiu, G.~Wen, Y.~Fan, Occluded object detection in high-resolution remote
  sensing images using partial configuration object model, IEEE Journal of
  Selected Topics in Applied Earth Observations and Remote Sensing 10~(5)
  (2017) 1909--1925.

\bibitem{qiu2017automatic}
S.~Qiu, G.~Wen, Z.~Deng, Y.~Fan, B.~Hui, Automatic and fast pcm generation for
  occluded object detection in high-resolution remote sensing images, IEEE
  Geosci. Remote Sens. Lett 14 (2017) 1730--1734.

\bibitem{qiu2018unified}
S.~Qiu, G.~Wen, J.~Liu, Z.~Deng, Y.~Fan, Unified partial configuration model
  framework for fast partially occluded object detection in high-resolution
  remote sensing images, Remote Sensing 10~(3) (2018) 464.

\bibitem{ding2018learning}
J.~Ding, N.~Xue, Y.~Long, G.-S. Xia, Q.~Lu, Learning roi transformer for
  detecting oriented objects in aerial images, arXiv preprint arXiv:1812.00155
  (2018).

\bibitem{zhang2019geospatial}
X.~Zhang, K.~Zhu, G.~Chen, X.~Tan, L.~Zhang, F.~Dai, P.~Liao, Y.~Gong,
  Geospatial object detection on high resolution remote sensing imagery based
  on double multi-scale feature pyramid network, Remote Sensing 11~(7) (2019)
  755.

\end{thebibliography}

\end{document}